\documentclass{article}

\usepackage{microtype}
\usepackage{graphicx}
\usepackage{subcaption} 
\usepackage{booktabs} 

\usepackage[hyphens]{url}
\usepackage{hyperref}
\hypersetup{breaklinks=true}

\usepackage{relsize}
\usepackage{array}
\usepackage{longtable} 
\usepackage{rotating}  

\usepackage{amssymb} 
\usepackage{enumitem}

\usepackage{tikz} 

\newcommand{\reqsym}{
  \tikz[baseline=-0.5ex]\filldraw[black] (0,0) circle (0.7ex);%
}
\newcommand{\optsym}{
  \tikz[baseline=-0.5ex]{%
    \draw[black] (0,0) circle (0.7ex);
    \filldraw[black] (0,0) circle (0.25ex); 
  }%
}
\newcommand{\helpsym}{
  \tikz[baseline=-0.5ex]\draw[black] (0,0) circle (0.7ex);%
}   

\usepackage[accepted]{icml2025}

\usepackage{amsmath}
\usepackage{amssymb}
\usepackage{mathtools}
\usepackage{amsthm}
\usepackage{multicol}

\usepackage[capitalize,noabbrev]{cleveref}

\setlist[itemize]{leftmargin=*}

\usepackage{longtable} 
\usepackage{rotating}  
\usepackage{url}       

\newcolumntype{P}[1]{>{\raggedright\arraybackslash}p{#1}}
\newcolumntype{Y}{>{\centering\arraybackslash}p{2.5ex}}

\theoremstyle{plain}

\theoremstyle{definition}

\theoremstyle{remark}

\usepackage[textsize=tiny]{todonotes}

\icmltitlerunning{Technical Requirements for Halting Dangerous AI Activities} 

\begin{document}

\twocolumn[ 
\icmltitle{Technical Requirements for Halting Dangerous AI Activities}

\icmlsetsymbol{equal}{*}

\begin{icmlauthorlist}
\icmlauthor{Peter Barnett}{comp}
\icmlauthor{Aaron Scher}{comp}
\icmlauthor{David Abecassis}{comp}
\end{icmlauthorlist}

\icmlaffiliation{comp}{Machine Intelligence Research Institute, CA, USA} 

\icmlcorrespondingauthor{Peter Barnett}{peter@intelligence.org} 

\icmlkeywords{Artificial Intelligence, AI Safety, AI Governance, Risk Management, AI Off Switch, Compute Governance}

\vskip 0.3in
] 

\printAffiliationsAndNotice{} 

\begin{abstract}
The rapid development of AI systems poses unprecedented risks, including loss of control, misuse, geopolitical instability, and concentration of power.  To navigate these risks and avoid worst-case outcomes, governments may proactively establish the capability for a coordinated halt on dangerous AI development and deployment. In this paper, we outline key technical interventions that could allow for a coordinated halt on dangerous AI activities. We discuss how these interventions may contribute to restricting various dangerous AI activities, and show how these interventions can form the technical foundation for potential AI governance plans.  
\end{abstract}

\vspace{-2.75ex}
\section{Introduction} 
Humanity appears to be on a trajectory toward developing AI systems that substantially outperform human experts across many cognitive domains. While potentially beneficial, the development of advanced AI carries profound risks~\citep{cais_statement_2023, bengio2024managing}. These risks are exacerbated if the transition to advanced AI is very fast, as the world would have less time to prepare. A key driver may be AI systems that are capable of automating AI R\&D, leading to a feedback loop that rapidly increases AI capabilities. 

To avoid these risks and have time to prepare, governments may wish to coordinate to halt or restrict dangerous AI development and deployment~\citep{scholefield2025international}. Such a halt would address the following key risks:
\begin{itemize}[nosep]
    \item \textbf{Loss of Control:} AIs pursuing unintended goals could disempower humanity, potentially leading to extinction.
    \item \textbf{Misuse:} Malicious actors could use powerful AI for catastrophic ends~\citep{zelikow2024defense}, e.g., designing novel bioweapons~\citep{crawford2024securing} or orchestrating large-scale cyberattacks on critical infrastructure.
    \item \textbf{Geopolitical Instability and War:} An AI arms race or the potential emergence of a strategically dominant AI could trigger conflicts.
    \item \textbf{Concentration of Power:} AI could lead to an extreme concentration of power, potentially enabling authoritarian lock-in through unprecedented surveillance and control~\cite{davidson2025aienabledcoupsho}.
\end{itemize}
Beyond mitigating these risks, the design of a halt must also consider political viability, the preservation of AI's benefits, and reversibility once safety is assured.
\section{Halting or Restricting AI Activities}
The main goal of a halt is to stop the advancement of AI capabilities globally, but the mechanisms used in a halt will also be useful for verifiably restricting or halting specific, dangerous AI activities. Based on the current trajectory of AI development, we believe the main intervention point for these mechanisms is \emph{AI compute}~\cite{sastry_computing_2024}. Plans generally revolve around limiting access to large quantities of advanced AI chips or monitoring the use of these chips, including by shutting them off or monitoring workloads~\citep{scher_mechanisms_2024}. Other than compute, there are various approaches to \emph{monitoring AI activities}, such as through mandatory reporting requirements~\citep{belfield_what_2024}, auditing, and espionage. Another key function of halt mechanisms is \emph{limiting proliferation} of dangerous AI capabilities. Besides compute, AI progress is largely driven algorithmic advances~\citep{ho_algorithmic_2024}. Depending on the state of AI capabilities, it may also be necessary to \emph{restrict algorithmic progress}. 

Authorities will require different capacities, depending on the risk landscape. These capacities may include: \textbf{restricting training} (e.g., via compute thresholds), \textbf{restricting inference} (e.g., preventing misuse or uncontrolled automated AI R\&D), and \textbf{restricting post-training}. These three capacities are useful for understanding which part of AI development a particular intervention helps address. For instance, if dangerous AI systems have already been trained, governance would need to focus on inference, and training-specific interventions would be less relevant. Capacities help organize the space of interventions, but our primary analysis concerns AI governance plans (\cref{sec:plans} and which interventions are useful or necessary for those plans. In this paper, we primarily discuss non-emergency technical interventions; see \cref{app:emergency_shut_down} for emergency shutdown interventions.

\section{Technical Interventions}
\label{sec:technical_interventions}
Halting or restricting AI activities requires a suite of interconnected technical capabilities, likely including some subset of the following interventions.

\subsection{Chip location} 
Tracking the location of AI hardware is essential for many compute governance approaches; authorities must know the location of AI hardware to inspect it or shut it down.
\begin{itemize}[nosep]
    \item \textbf{Track chip shipments via manufacturers/distributors}: Engage suppliers to track AI chip distribution and deployment locations, covering both historical, ongoing, and future shipments.
    \item \textbf{Hardware-enabled location tracking}: Use secure hardware features (e.g., FlexHEGs~\citep{petrie2024interim}) for chips to verify their location and status remotely~\citep{brass2024location, aarne2024secure}.
    \item \textbf{Centralize compute in declared datacenters}: Consolidate AI compute in a small number of secure, registered datacenters to simplify tracking and control. 
    \item \textbf{Periodic datacenter inspections}: Conduct periodic physical audits of datacenters to verify chip inventories.
    \item \textbf{Ongoing datacenter monitoring}: Establish continuous governmental oversight of datacenters, which may leverage inspectors or tamper-resistant surveillance cameras.
\end{itemize}

\subsection{Chip manufacturing} 
Interventions targeting semiconductor manufacturing offer leverage at an earlier stage of AI development. Preventing certain actors from gaining access to hardware could effectively block them from doing dangerous AI activities.
\begin{itemize}[nosep]
    \item \textbf{Monitoring for the construction of new fabs}: Surveil industrial development to detect the construction of new advanced AI chip fabrication plants~\citep{wasil2024verification}.
    \item \textbf{Restrict/control the equipment and materials}: Restrict access to critical equipment (e.g., EUV machines) and material inputs (e.g., silicon wafers) for advanced chip production~\citep{thadani2023mapping}.
    \item \textbf{Surveillance and inspections of fabs}: Conduct surveillance and on-site inspections of semiconductor fabs to verify compliance with agreements or restrictions. For example, only producing a maximum number of chips or only manufacturing authorized chip designs.
    \item \textbf{Verifiably shut down fabs}: Deactivate specific chip fabs, verifiable via satellite imagery, electricity monitoring, etc.
    \item \textbf{Verify that new chips have HEMs to spec}: Mandate and verify that new AI chips have hardware-enabled governance mechanisms (HEMs)~\citep{kulp2024hardwareenabled} (e.g., FlexHEGs)~\citep{petrie2024interim}.
\end{itemize}

\subsection{Compute monitoring} 
Beyond tracking hardware, controlling its actual use is paramount. This involves establishing thresholds and implementing mechanisms to monitor and regulate how AI compute resources are employed.
\begin{itemize}[nosep]
    \item \textbf{Define capabilities thresholds}: Define measurable AI capability levels (e.g., autonomous replication) that trigger specific required responses (e.g., halt development, implement security measures)~\citep{karnofsky2024sketch, koessler2024risk, shevlane_model_2023, raman2025intolerable}.
    \item \textbf{Define compute thresholds}: Establish limits on compute used for training models~\citep{heim_training_2024}.
    \item \textbf{Determine which datacenters need monitoring}: Establish criteria for datacenters that require oversight based on their compute capacity. 
    \item \textbf{Monitor or restrict at the datacenter level}: Implement controls directly at datacenters, such as chip kill-switches, interconnect limits, workload classification based on external measurements~\citep{scher_mechanisms_2024}.
    \item \textbf{Monitor or restrict via hardware-enabled mechanisms}: Leverage specialized hardware features to enforce usage policies or monitor AI activities~\citep{kulp2024hardwareenabled, aarne2024secure, petrie2024interim}. 
    \item \textbf{Monitor or restrict via software}: Employ software-based tools for monitoring AI tasks, analyzing code executing on compute, or restricting certain operations~\citep{scher_mechanisms_2024}. This may be easier to subvert than other methods.
    \item \textbf{Monitor or restrict consumer compute}: Limit sales of consumer compute (e.g., personal computers, gaming GPUs, etc), and potentially mandate monitoring for certain hardware.
    \item \textbf{Inference-only hardware}: Develop AI hardware architecturally limited to inference (i.e., incapable of efficient model training) to prevent prohibited development.
\end{itemize}

\subsection{Non-compute monitoring} 
Monitoring AI models and AI development projects, rather than their compute resources. These interventions would provide increased transparency into AI development.
\begin{itemize}[nosep]
    \item \textbf{Required reporting of capabilities}: Implement mandatory disclosure  of model capabilities~\citep{belfield_what_2024}.
    \item \textbf{Third-party/government evaluations of AI models}: Establish processes for independent AI evaluations by third-parties~\citep{casper_black-box_2024}.
    \item \textbf{Other inspections for safety/security practices}: Conduct broader inspections of AI development facilities focusing on safety and security practices~\citep{stix2025ai}.
    \item \textbf{In-house auditors}: Mandate AI developers have embedded audit teams for capability assessment and compliance, similar to ``compliance monitors'' in the finance industry.
    \item \textbf{Automated auditors}: Use AI systems to monitor AI development. These automated auditors could catch prohibited activities while preserving privacy.
    \item \textbf{Espionage}: Governments could use existing espionage tactics to gather information on AI development.
    \item \textbf{Whistleblowers}: Establish protected channels for insiders to report concerns about dangerous AI development. 
\end{itemize}

\subsection{Limiting proliferation} 
If powerful AI model weights or critical algorithmic insights are made available to malicious actors or the public, preventing their use becomes very challenging~\citep{seger2023open}. Authorities could prevent the proliferation of models by preventing them from being trained in the first place. 
\begin{itemize}[nosep]
    \item \textbf{Model weight security}: Implement and verify security to protect AI model weights from unauthorized access or leakage~\cite{nevo_securing_2024}.
    \item \textbf{Algorithmic secret security}: Protect algorithmic insights from proliferation.
    \item \textbf{Structured access}: Rather than providing broad access to potentially dangerous AI systems, provide controlled access to a limited set of users~\cite{shevlane2022structured}.
    \item \textbf{Limit open model release}: Implement policies to restrict the broad release of AI models with high-risk capabilities. 
    \item \textbf{Hardware-specific models}: Develop AI models which can only run on specific hardware~\citep{clifford2024locking}. If model weights leak, the model cannot easily be run without the required hardware.
    \item \textbf{Non-fine-tunable models}: Create methods for training models that cannot be fine-tuned to unlock new dangerous capabilities~\citep{deng2024sophon, tamirisa2024tamper}. 
\end{itemize}

\subsection{Research} 
Unmonitored compute would pose risks if algorithmic progress continues~\citep{ho_algorithmic_2024}. 
Oversight could help prohibit dangerous research while enabling beneficial work.
\begin{itemize}[nosep]
    \item \textbf{Tracking personnel}: Track activities of key researchers. This could include tracking their location and affiliation, to ensure they are not part of a secret AI development project.
    \item \textbf{Define and prohibit algorithmic progress research}: Identify and ban specific lines of algorithmic research deemed too dangerous or destabilizing. 
    \item \textbf{Surveillance of AI research activities}: Monitor the computers and research activities of AI researchers to ensure compliance with research guidelines.
    \item \textbf{Model spec prohibiting algorithmic research}: Include prohibitions on accelerating AI research in AI model specifications~\citep{openai_openai_2025, scher_mechanisms_2024}. 
\end{itemize}

\section{Plans}
\label{sec:plans}
These technical interventions form the basis for various proposed AI governance plans, summarized in this section. \cref{tab:offswitch_updated_may8} shows the technical requirements for each plan.
\paragraph{Last-minute Wake-up}
World leaders may not take substantial steps to halt AI development until AI capabilities are very advanced and substantial harm has occurred. The Last-minute Wake-up plan attempts to halt AI development in that world, first responding to an acute emergency (\cref{app:emergency_shut_down}) and then implementing a global compute monitoring regime. Depending on progress in the development of verification mechanisms and methods for predicting risk, effective compute monitoring regimes differ substantially in their details. For instance, compute monitoring could involve completely shutting off chips or permitting some verified workloads~\citep{scher_mechanisms_2024}.
Verification could allow society to safely benefit from  AI systems.  
This regime would primarily prevent the advancement of AI capabilities via strict limits on the amount of compute used for AI training. It may also need to monitor or restrict the post-training and inference of existing AI models, if they pose a risk. Hopefully, consumer computers would be exempt from this compute monitoring regime, unless sufficiently advanced AI models have proliferated or the compute requirements for dangerous AI activities are very low (e.g., due to substantial algorithmic progress). Therefore, this plan also involves slowing the rate of algorithmic progress by banning relevant research and avoiding the proliferation of the most capable models via internal security measures and limitations on model release.
  
\paragraph{Chip Production Moratorium}
The compute hardware required to enable dangerous AI may not yet have been built. The amount of AI-relevant compute is expected to increase tenfold in the next two years~\citep{dean2025compute}, and hundredfold by 2030~\citep{pilz2025trends, epoch2024canaiscalingcontinuethrough2030}. 
Progress in the near-term is likely to rely on continued scaling of this input, production of which is concentrated in dozens of highly advanced chip fabs~\citep{sastry_computing_2024}.

This moratorium aims to globally pause the production of new AI compute. This is made feasible due to the cost, size, and fragility of semiconductor production. It avoids difficulties common to AI verification by requiring only that new cutting-edge chips are not produced, which rival states can independently verify with ease. As a last resort, it may even be implemented unilaterally, though in practice, the possibility of unilateral action may promote coordination between rivals. This plan's effectiveness depends on how soon it can be enacted. Governing existing compute and restricting algorithmic research are helpful extensions, but they may not be necessary to delay dangerous AI by decades.

\paragraph{A Narrow Path}
\citet{miotti_narrow_2024} proposes immediate national actions and an international treaty to prevent the development of artificial superintelligence (ASI) for two decades. It establishes a ``defense in depth'' strategy using regulatory oversight, including a licensing regime based on compute thresholds, prohibitions on dangerous capabilities like AI self-improvement, and mandatory safety justifications for AI systems.
\paragraph{Keep the Future Human}
\citet{aguirre2025keep} advocates preventing the development of smarter-than-human, autonomous, general-purpose AI---and ASI---by focusing on hardware-enabled compute governance to enforce limits on the compute used to create AI systems. It also manages risk through enhanced liability when AI systems feature autonomy, generality, and intelligence.
\paragraph{Superintelligence Strategy}
\citet{hendrycks_superintelligence_2025} proposes a national security framework based on deterrence via Mutual Assured AI Malfunction (MAIM). States use threats of sabotage to prevent each other from developing destabilizing AI capabilities. This is combined with nonproliferation efforts against rogue actors, such as hardware export control, algorithmic secret security, and model weight security. 

\section{Assessment of Plans and Requirements}
In \cref{tab:offswitch_updated_may8}, we provide an overview of the cataloged interventions, and the relevance of each intervention to the key capacities (restricting training, inference, and post-training) and plans.

The plans in \cref{sec:plans}were highlighted because they, to some extent, involve halting dangerous AI activities. The prior plans (\citet{miotti_narrow_2024}, \citet{aguirre2025keep}, \citet{hendrycks_superintelligence_2025}) are some of the only plans for a halt described in sufficient detail to allow assessment.
The novel plans (Last-minute Wake-up, and Chip Production Moratorium) are intended to substantially reduce catastrophic risk from advanced AI. The plans were also chosen to be diverse, covering different approaches to managing catastrophic AI risks. 

We grade the technological readiness of each of the interventions as follows:
\begin{itemize}[nosep]
    \item \textbf{High}: No technological hurdles to implement this intervention.
    \item \textbf{Medium}: Significant effort needed for real-world implementation.
    \item \textbf{Low}:  Lacks even functional prototypes, important aspects have not yet been defined.
\end{itemize}

For each intervention and each plan, we assess the extent to which the plan depends on the intervention.  \emph{Required} means that this intervention is essential to a given plan or capability. \emph{Maybe required} means that this intervention may be needed; for example if there are multiple ways to fulfill a specific function (e.g., methods to oversee AI research) or if there is substantial uncertainty about whether the function will be needed under a plan (e.g., some interventions are only important if AI capabilities have already advanced to dangerous levels, which may or may not happen before halting). \emph{Helpful} means that this intervention is useful for some functions of a plan, but not essential. 

These assessments are preliminary best estimates and subject to refinement. For previously published plans, assessments are based on identifying the functions that plans aim to accomplish, whether the plan explicitly mentions the intervention, and whether we think the plan implicitly relies on the intervention. For plans first discussed here, we reach this assessment by identifying the main functions the plans rely on and key interventions to perform these functions.

\section{Conclusion}
Our analysis of technical requirements for halting dangerous AI activities can improve understanding and prioritization of AI governance work. All of these plans requires substantial
control over AI compute through chip tracking, datacenter monitoring, or manufacturing restrictions. For most plans, limiting model access via security measures and restricting open release is essential, as proliferation could make control of models impossible. Most plans also rely on interventions currently at low technological readiness.

These findings highlight urgent priorities. The required infrastructure and technology must be developed before it is needed, such as hardware-enabled mechanisms.  International tracking of AI hardware should begin soon, as this is crucial for many plans and will only become more
difficult if delayed. Without significant effort now, it will be difficult to halt in the future, even if there is will to do so.

\newpage
\onecolumn 
\begin{footnotesize} 
\setlength{\tabcolsep}{3pt} 
\begin{longtable}{@{} p{0.4\linewidth} c || *{3}{Y} || *{5}{Y} @{}}

\caption{Technical interventions for halting or restricting dangerous AI activities, and the capacities and plans they are required for.} \label{tab:offswitch_updated_may8}\vspace{-0.3cm}
\\
\toprule

& & \multicolumn{3}{c||}
{\textbf{Capacities}} & \multicolumn{5}{c}{\textbf{Plans}} \\
\cmidrule(lr){3-5} \cmidrule(lr){6-10}
\multicolumn{1}{l}{
  \begin{minipage}[t]{8cm}
    \raggedright 
    \setlength{\fboxsep}{4pt}
    \setlength{\fboxrule}{0.4pt}
    \fbox{\begin{tabular}{@{}l@{}} 
        \textbf{Legend:} \\
      \reqsym~=~Required \\
      \optsym~=~Maybe required, depending on the situation \\
      \helpsym~=~Helpful, but not required
    \end{tabular}}%
    \par 
    \vspace{4.5ex} 
    \textbf{Intervention}%
    \vspace{1ex}
  \end{minipage}%
} &
\rotatebox[origin=c]{90}{\textbf{Technological Readiness}} &
\rotatebox[origin=c]{90}{\textbf{Restrict Training}} &
\rotatebox[origin=c]{90}{\textbf{Restrict Inference}} &
\rotatebox[origin=c]{90}{\textbf{Restrict Post-training}} &
\rotatebox[origin=c]{90}{\textbf{Last-minute Wake-up}} & 
\rotatebox[origin=c]{90}{\textbf{Chip Production Moratorium}} &                 
\rotatebox[origin=c]{90}{\textbf{A Narrow Path$^*$}} &
\rotatebox[origin=c]{90}{\textbf{Keep the Future Human$^*$}} &
\rotatebox[origin=c]{90}{\textbf{Superintelligence Strategy$^*$}} \\
\midrule
\endfirsthead

\caption*{Table: AI Off Switch - Technical Requirements \& Functions (Continued)} \\
\toprule
& & \multicolumn{3}{c||}{\textbf{Capacities}} & \multicolumn{5}{c}{\textbf{Plans}} \\
\cmidrule(lr){3-5} \cmidrule(lr){6-10}
\multicolumn{1}{l}{
  \begin{minipage}[t]{8cm}
    \raggedright 
    \setlength{\fboxsep}{4pt}
    \setlength{\fboxrule}{0.4pt}
    \fbox{\begin{tabular}{@{}l@{}} 
      \reqsym~=~Required \\
      \optsym~=~Maybe required, depending on the situation \\
      \helpsym~=~Helpful but not required
    \end{tabular}}%
    \par 
    \vspace{1.5ex} 
    \textbf{Intervention}%
  \end{minipage}%
} &
\rotatebox[origin=c]{90}{\textbf{Tech. Readiness}} &
\rotatebox[origin=c]{90}{\textbf{Restrict Training}} &
\rotatebox[origin=c]{90}{\textbf{Restrict Inference}} &
\rotatebox[origin=c]{90}{\textbf{Restrict Post-training}} &
\rotatebox[origin=c]{90}{\textbf{Last-minute wake-up}} & 
\rotatebox[origin=c]{90}{\textbf{MSLS}} &               
\rotatebox[origin=c]{90}{\textbf{A Narrow Path}} &
\rotatebox[origin=c]{90}{\textbf{Keep the Future Human}} &
\rotatebox[origin=c]{90}{\textbf{Superintelligence Strategy}} \\
\midrule
\endhead

\midrule 
\multicolumn{10}{l}{{\fontsize{8.4}{8.4} \selectfont$^*$This appraisal of the technical requirements for these plans is not necessarily endorsed by their original authors.}} \\
\bottomrule 
\endlastfoot

\addlinespace[0.3em]
\multicolumn{10}{@{}l}{\textbf{Chip location}} \\
\midrule
Track chip shipments via manufacturers/distributors & High & \optsym & \helpsym & \helpsym & \reqsym & \optsym & \reqsym & \reqsym & \reqsym \\ %
Hardware-enabled location tracking & Low & \helpsym & \helpsym & \helpsym & \helpsym & \optsym & \optsym & \helpsym & \optsym \\ %
Centralize compute in declared datacenters & Medium & \helpsym & \helpsym & \helpsym & \optsym & \optsym & \helpsym & \helpsym & \helpsym \\ %
Periodic datacenter inspections & High & \optsym & \optsym &  & \reqsym & \helpsym & \optsym & \helpsym & \helpsym \\ %
Ongoing datacenter monitoring & High & \optsym & \optsym & \helpsym & \reqsym & \helpsym & \helpsym & \helpsym & \reqsym \\ %
\midrule
\addlinespace[0.3em]
\multicolumn{10}{@{}l}{\textbf{Chip manufacturing}} \\
\midrule
Monitoring for the construction of new fabs & High & \optsym &  &  & \reqsym & \reqsym & \helpsym & \reqsym & \helpsym \\ %
Restrict/control the equipment and materials & Medium & \optsym &  &  & \helpsym & \optsym & \helpsym & \helpsym & \helpsym \\ %
Surveillance and inspections of fabs & Medium & \helpsym &  &  & \reqsym & \reqsym &  & \helpsym & \helpsym \\ %
Verifiably shut down fabs & High & \optsym &  &  &  & \reqsym &  & \helpsym & \helpsym \\ %
Verify that new chips have HEMs to spec & Low & \helpsym & \optsym & \optsym & \helpsym &  & \helpsym & \reqsym & \helpsym \\ %
\midrule
\addlinespace[0.3em]
\multicolumn{10}{@{}l}{\textbf{Compute monitoring}} \\
\midrule
Define capabilities thresholds & Medium & \optsym & \reqsym & \reqsym & \reqsym & \optsym & \reqsym & \reqsym & \reqsym \\ %
Define compute thresholds & High & \optsym &  & \helpsym & \reqsym & \optsym & \reqsym & \reqsym & \helpsym \\ %
Determine which datacenters need monitoring & Medium & \reqsym & \optsym &  & \reqsym & \optsym & \reqsym & \reqsym & \reqsym \\ %
Monitor or restrict at the datacenter level & High & \optsym & \optsym & \helpsym & \reqsym &  & \reqsym & \optsym & \helpsym \\ %
Monitor or restrict via hardware-enabled mechanisms & Low & \optsym & \optsym & \helpsym & \helpsym &  & \helpsym & \reqsym & \helpsym \\ %
Monitor or restrict via software & Low & \helpsym & \optsym & \helpsym & \helpsym &  & \helpsym & \helpsym & \helpsym \\ %
Inference-only hardware & Medium & \optsym &  & \optsym & \helpsym & \helpsym &  & \helpsym & \helpsym \\ %
Monitor or restrict consumer compute & Low & \optsym & \optsym & \optsym & \optsym &  &  &  &  \\ %
\midrule
\addlinespace[0.3em]
\multicolumn{10}{@{}l}{\textbf{Non-compute monitoring}} \\
\midrule
Required reporting of capabilities & High &  & \reqsym & \reqsym & \reqsym &  & \reqsym & \reqsym & \reqsym \\ %
Third-party/government evaluations of AI models & High & \optsym & \optsym & \helpsym & \optsym &  & \reqsym & \reqsym & \reqsym \\ %
Other inspections for safety/security practices & Medium & \helpsym &  &  & \optsym &  & \reqsym & \reqsym & \helpsym \\ %
In-house auditors & Medium & \optsym & \optsym & \optsym & \optsym &  & \helpsym & \helpsym & \helpsym \\ %
Automated auditors & Low & \optsym & \helpsym & \optsym & \optsym &  &  & \helpsym & \helpsym \\ %
Espionage & High & \optsym & \helpsym & \optsym & \reqsym & \helpsym &  &  & \reqsym \\ %
Whistleblowers & High & \optsym & \optsym & \optsym & \helpsym & \helpsym & \helpsym & \helpsym & \helpsym \\ %
\midrule
\addlinespace[0.3em]
\multicolumn{10}{@{}l}{\textbf{Limiting proliferation}} \\
\midrule
Model weight security & Medium & \optsym & \reqsym & \reqsym & \reqsym &  & \reqsym & \reqsym & \reqsym \\ %
Algorithmic secret security & Medium & \optsym &  & \helpsym & \optsym & \helpsym & \helpsym & \helpsym & \reqsym \\ %
Structured access & High & \optsym & \optsym & \optsym & \reqsym &  & \reqsym & \helpsym & \reqsym \\ %
Limit open model release & High & \helpsym & \reqsym & \reqsym & \reqsym &  & \reqsym & \reqsym & \reqsym \\ %
Hardware-specific models & Low & \helpsym & \optsym & \helpsym & \helpsym &  &  & \helpsym & \helpsym \\ %
Non-fine-tunable models & Low &  &  & \helpsym & \helpsym &  & \helpsym & \helpsym & \helpsym \\ %
\midrule
\addlinespace[0.3em]
\multicolumn{10}{@{}l}{\textbf{Research}} \\
\midrule
Tracking personnel & High & \helpsym & \helpsym &  & \optsym & \helpsym &  &  & \optsym \\ %
Define and prohibit algorithmic progress research & Medium & \helpsym & \optsym &  & \optsym & \optsym & \reqsym & \optsym & \reqsym \\ 
Surveillance of AI research activities & Medium & \helpsym & \optsym & \helpsym & \optsym & \optsym & \reqsym &  & \optsym \\ %
Model spec prohibiting algorithmic research & High & \helpsym & \optsym &  & \optsym & \helpsym & \helpsym & \optsym & \helpsym \\ 
\end{longtable}
\end{footnotesize}

\twocolumn 

\section*{Impact Statement}
We try to advance the understanding of technical requirements for governing potentially dangerous AI systems. The development of capabilities for halting or restricting dangerous AI activities, while intended to mitigate catastrophic risks, could itself have complex societal consequences.

These measures, targeted at reducing catastrophic risk, represent significant expansion of international governmental authority over AI development. Regulating dangerous AI after it is developed and deployed will require much more invasive measures, and may not be possible at all. Therefore, targeted oversight of AI activities and key inputs could serve to minimize future government intrusion while averting large-scale harm. 

We acknowledge that restricting certain AI activities would necessarily slow the realization of some technological benefits. However, we believe that many advantages of narrow AI systems remain accessible under these regimes, while avoiding the most destabilizing and dangerous effects of unrestricted development. The optimal approach prioritizes careful progress rather than unrestricted advancement, allowing research to move forward when it is safe.

%
\bibliography{references}
\bibliographystyle{icml2025}

\appendix
\section{Emergency Shutdown}
\label{app:emergency_shut_down}
The body of this short paper has focused on technical interventions to halt or restrict dangerous AI activities in non-emergency situations. However, it may be crucial for authorities to have emergency shutdown capacities, capacities that allow them to rapidly shut down and limit the harm from dangerous AI systems. For example, this could include shutting down a datacenter running a rogue autonomous AI or an AI being used for a large-scale cyberattack by a malicious actor~\citep{the-rogue-replication-threat-model, shlegeris2024aicatastrophes}. An emergency may, unfortunately, be necessary for governments to become sufficiently concerned about dangerous AI, such that they consider plans to halt or restrict dangerous AI activities. 

Various technical interventions previously mentioned in \cref{sec:technical_interventions} would be useful in an emergency shutdown situation, specifically measures around chip location, compute use, and limiting proliferation. Here we provide an initial list of technical interventions specifically useful for emergency shutdown. 
\begin{itemize}
    \item \textbf{Track down rogue AIs}: Find out which physical hardware a rogue AI system is running on, in order to shut it off. An example would be tracking an AI system via its IP address. In an optimistic scenario, a rogue AI system will have only spread to a limited number of datacenters that are easily located and shut off. A more pessimistic scenario could involve many copies of an AI system running on consumer hardware or running on a distributed botnet. Tools like AI watermarking and AI text detection may help identify and locate rogue AI systems. Tracking down rogue AIs could be very difficult or effectively impossible.
    \item \textbf{AI spread reduction}: Prevent rogue AI systems spreading and accumulating resources~\citep{cohen_here_2025}. This could include strong know-your-customer (KYC) or proof-of-humanity verification on services a rogue AI might use, such as cloud compute providers, online financial services, and online work platforms like Upwork~\citep{chan2025infrastructure}. Other interventions include wide-scale preemptive red-teaming of cyber vulnerabilities or publicity campaigns for humans to not cooperate with rogue AI systems.
    \item \textbf{Last-resort shutdown measures}: If AI systems are contained to datacenters, shutting them down is relatively technically straightforward. It would be much more difficult to shut down an AI system that has spread widely across non-datacenter compute (such as personal computers). It is unclear if shutting down such a system is technically feasible, but approaches for further research might include shutting down parts of the electrical grid, EMP weapons to disable electronics in a wide area, and pervasive cyber weapons.  
    \item \textbf{Circuit breakers on societal infrastructure}: AI systems may cause cascading failures in complex systems such as the stock market, power grid, or supply chains. Such systems should have fail-safe mechanisms and other components to mitigate such failures. 
\end{itemize}

\end{document}